\documentclass[10pt]{article}

\usepackage[utf8]{inputenc}
\usepackage[T1]{fontenc}

\usepackage{epsf}
\usepackage{amsmath}

\allowdisplaybreaks

\usepackage[showframe=false]{geometry}
\usepackage{changepage}

\usepackage{epsfig}
\usepackage{amssymb}

\usepackage{amsthm}
\usepackage{setspace}
\usepackage{cite}
\usepackage{mcite}

\usepackage{algorithmic}  
\usepackage{algorithm}

\usepackage{shadow}
\usepackage{fancybox}
\usepackage{fancyhdr}

\usepackage{color}
\usepackage[usenames,dvipsnames,svgnames,table]{xcolor}
\newcommand{\bl}[1]{\textcolor{blue}{#1}}
\newcommand{\red}[1]{\textcolor{red}{#1}}

\definecolor{mypurple}{rgb}{.4,.0,.5}
\newcommand{\prp}[1]{\textcolor{mypurple}{#1}}

\usepackage[hyphens]{url}

\usepackage[colorlinks=true,
            linkcolor=black,
            urlcolor=blue,
            citecolor=purple]{hyperref}

\usepackage{breakurl}

\def\y{{\bf y}}

\def\x{{\bf x}}

\def\x{{\mathbf x}}

\def\x{{\bf x}}
\def\y{{\bf y}}
\def\z{{\bf z}}

\def\a{{\bf a}}
\def\b{{\bf b}}

\def\d{{\bf d}}
\def\f{{\bf f}}

\def\tr{\mbox{Tr}}

\def\tr{{\rm tr}\,}

\def\cS{{\mathcal S}}

\def\be{\begin{equation}}
\def\ee{\end{equation}}
\def\ba{\left[\begin{array}}
\def\ea{\end{array}\right]}

\def\x{{\bf x}}
\def\y{{\bf y}}
\def\z{{\bf z}}

\def\a{{\bf a}}
\def\b{{\bf b}}

\def\d{{\bf d}}
\def\f{{\bf f}}

\def\1{{\bf 1}}

\def\g{{\bf g}}
\def\0{{\bf 0}}

\def\mR{{\mathbb R}}

\def\mE{{\mathbb E}}

\def\mP{{\mathbb P}}

\def\lp{\left (}
\def\rp{\right )}

\def\y{{\bf y}}

\def\x{{\bf x}}

\def\x{{\mathbf x}}

\def\x{{\bf x}}
\def\y{{\bf y}}
\def\z{{\bf z}}

\def\a{{\bf a}}
\def\b{{\bf b}}

\def\d{{\bf d}}
\def\f{{\bf f}}

\def\tr{\mbox{Tr}}

\def\tr{{\rm tr}\,}

\def\be{\begin{equation}}
\def\ee{\end{equation}}
\def\ba{\left[\begin{array}}
\def\ea{\end{array}\right]}

\def\x{{\bf x}}
\def\y{{\bf y}}
\def\z{{\bf z}}

\def\a{{\bf a}}
\def\b{{\bf b}}

\def\d{{\bf d}}
\def\f{{\bf f}}

\def\({\left (}
\def\){\right )}

\def\1{{\bf 1}}

\def\g{{\bf g}}
\def\0{{\bf 0}}

\usepackage{xcolor}
\usepackage{color}

\definecolor{darkgreen}{rgb}{0, 0.4,0}
\newcommand{\dgr}[1]{\textcolor{darkgreen}{#1}}

\definecolor{purplebrown}{rgb}{0.5,0.1,0.6}

\definecolor{ultclupcol}{rgb}{0.1,0.5,0.5}

\definecolor{mytrycolor}{rgb}{0.5,0.7,0.2}

\definecolor{ultclupcola}{rgb}{.5,0,.5}

\definecolor{shadebrown}{rgb}{0.1,0.1,0.9}
\definecolor{lightblue}{rgb}{0.2,0,1}

\usepackage{fancybox}
\usepackage{graphicx}
\usepackage{epstopdf}
\usepackage{epsfig}
\usepackage{wrapfig}
\usepackage{subfigure}

\usepackage{xcolor}
\usepackage{tcolorbox}
\tcbuselibrary{skins}

\newtcbox{\xmybox}{on line,
arc=7pt,
before upper={\rule[-3pt]{0pt}{10pt}},boxrule=0pt,
boxsep=0pt,left=6pt,right=6pt,top=0pt,bottom=0pt,enhanced, coltext=blue, colback=white!10!yellow}

\newtcbox{\xmyboxa}{on line,
arc=7pt,
before upper={\rule[-3pt]{0pt}{10pt}},boxrule=0pt,
boxsep=0pt,left=6pt,right=6pt,top=0pt,bottom=0pt,enhanced, colback=white!10!yellow}

\newtcbox{\xmyboxb}{on line,
arc=7pt,
before upper={\rule[-3pt]{0pt}{10pt}},boxrule=1pt,colframe=darkgreen!100!blue,
boxsep=0pt,left=6pt,right=6pt,top=0pt,bottom=0pt,enhanced, colback=white!10!yellow}

\newtcbox{\xmyboxc}{on line,
arc=7pt,
before upper={\rule[-3pt]{0pt}{10pt}},boxrule=.7pt,colframe=blue!100!blue,
boxsep=0pt,left=6pt,right=6pt,top=0pt,bottom=0pt,enhanced, coltext=blue, colback=white!10!yellow}

\newtcbox{\xmytboxa}{on line,
arc=7pt,
before upper={\rule[-3pt]{0pt}{10pt}},boxrule=.0pt,colframe=pink!50!yellow,
boxsep=0pt,left=6pt,right=6pt,top=0pt,bottom=0pt,enhanced, coltext=white, colback=blue!40!red}

\newtcbox{\xmytboxb}{on line,
arc=7pt,
before upper={\rule[-3pt]{0pt}{10pt}},boxrule=.0pt,colframe=pink!50!yellow,
boxsep=0pt,left=6pt,right=6pt,top=0pt,bottom=0pt,enhanced, coltext=white, colback=white!40!green}

\makeatletter
\newcommand\subsubsubsection{\@startsection{paragraph}{4}{\z@}{-2.5ex\@plus -1ex \@minus -.25ex}{1.25ex \@plus .25ex}{\normalfont\normalsize\bfseries}}
\newcommand\subsubsubsubsection{\@startsection{subparagraph}{5}{\z@}{-2.5ex\@plus -1ex \@minus -.25ex}{1.25ex \@plus .25ex}{\normalfont\normalsize\bfseries}}
\makeatother

\newtheorem{theorem}{Theorem}

\newtheorem{lemma}{Lemma}

\begin{document}

\begin{singlespace}

\title {\emph{Lifted} RDT based capacity analysis of the 1-hidden layer treelike \emph{sign} perceptrons neural networks 
}
\author{
\textsc{Mihailo Stojnic
\footnote{e-mail: {\tt flatoyer@gmail.com}} }}
\date{}
\maketitle

\centerline{{\bf Abstract}} \vspace*{0.1in}

We consider the memorization capabilities  of multilayered \emph{sign} perceptrons neural networks (SPNNs). A recent rigorous upper-bounding capacity characterization, obtained in \cite{Stojnictcmspnncaprdt23} utilizing the Random Duality Theory (RDT), demonstrated that adding neurons in a network configuration may indeed be very beneficial. Moreover, for particular \emph{treelike committee machines} (TCM)  architectures with $d\leq 5$ neurons in the hidden layer,  \cite{Stojnictcmspnncaprdt23} made a very first mathematically rigorous progress in over 30 years by lowering the previously best known capacity bounds of \cite{MitchDurb89}. Here, we first establish that the RDT bounds from \cite{Stojnictcmspnncaprdt23} scale as $\sim \sqrt{d}$ and can not on their own \emph{universally} (over the entire range of $d$) beat the best known $\sim \log(d)$ scaling of the bounds from \cite{MitchDurb89}. After recognizing that the progress from \cite{Stojnictcmspnncaprdt23} is therefore promising, but yet without a  complete concretization, we then proceed by considering the recently developed fully lifted RDT (fl RDT) as an alternative. While the fl RDT is indeed a powerful juggernaut, it typically relies on heavy numerical evaluations. To avoid such heavy numerics, we here focus on a simplified, \emph{partially lifted}, variant and show that it allows for very neat, closed form, analytical capacity characterizations. Moreover, we obtain the concrete capacity bounds that \emph{universally} improve for \emph{any} $d$ over the best known ones of \cite{MitchDurb89}.


\vspace*{0.25in} \noindent {\bf Index Terms: Multi-layer neural networks; Capacity; Lifted random duality theory}.

\end{singlespace}

\section{Introduction}
\label{sec:intro}

The last decade has seen an unprecedented level of demand for efficient collecting, interpreting, and managing of large data sets. Within such an environment, the machine learning (ML) concepts were quickly recognized as a tool that can be of great help. As a natural consequence, a rapid practical and theoretical development of various ML branches ensued, thereby dominating scientific approaches to big data handling for the larger portion of the last decade. Along the same lines, development of neural networks (NN) and, in particular, their algorithmic capabilities has been a concurrent process happening at a pace and scale never seen before. The great advancements made on the algorithmic front have closely been followed by the corresponding theoretical ones as well. In this paper we continue such a trend and analyze, from the theoretical point of view, one of the most fundamental NN features, the so-called, network's \emph{memory capacity}. To be able to properly explain what the memory capacity is and what the relevant problems of interests are as well as how we contribute to their a potential resolution, we below first recall on some of the basic NN models' properties.

\subsection{Neural networks basics (models, mathematical formalisms, prior work)}
\label{sec:model}

Multilayered multi-input single-output feed-forward neural nets with $L-2$ hidden layers and $d_i$ ($i\in\{1,2,\dots,L\}$) nodes (neurons) in the $i$-th layer will be the subject of our study in this paper. Layers $i=1$ and $i=L$ correspond to the network input and output, respectively. As such, they are somewhat artificial and will be referred to as layers to facilitate notational consistencies. Assuming that $\f^{(i)}(\cdot)=[\f_1^{(i)}(\cdot),\f_2^{(i)}(\cdot),\dots,\f_{d_{i+1}}^{(i)}(\cdot)]^T$ are the vectors of threshold functions $\f_{j}^{(i)}(\cdot):\mR^{d_i}\rightarrow \mR$ that describe how neuron $j$  in layer $i$ operates,
the network effectively functions by passing the outputs of the nodes from layer $i$ to $i+1$ through a linear combination governed by the matrix of weights $W^{(i)}\in\mR^{d_{i}\times d_{i+1}}$. Setting $\d=[d_1,d_2,\dots,d_L]$ (with $d_1=n$ and $d_L=1$) and denoting by $\x^{(i)}\in\mR^{d_i}$ and $\x^{(i+1)}\in\mR^{d_{i+1}}$ the inputs and outputs of neurons in layer $i$, one
for thresholds vectors $\b^{(i)}\in\mR^{d_{i+1}}$ and $i=1,2,\dots L$ has the following:
\begin{center}
     	\tcbset{beamer,nobeforeafter,lower separated=false, fonttitle=\bfseries, coltext=black,
		interior style={top color=yellow!20!white, bottom color=yellow!60!white},title style={left color=black, right color=red!50!blue!60!white},
		before=,after=\hfill,fonttitle=\bfseries,equal height group=AT}
\begin{tcolorbox}[title=Mathematical formalism of NN with architecture $A(\d\text{, }\f^{(i)})$:]
\vspace{-.03in}$\mbox{\textbf{input:}} \triangleq \x^{(1)}\quad \longrightarrow$ \hfill
\tcbox[coltext=black,colback=white!65!red!30!cyan,interior style={top color=yellow!20!white, bottom color=yellow!60!white},nobeforeafter,box align=base]{$ \x^{(i+1)}=\f^{(i)}(W^{(i)}\x^{(i)}-\b^{(i)}) $ }\hfill $\longrightarrow \quad \mbox{\textbf{output:}} \triangleq \x^{(L+1)}$.
\vspace{-.2in}\begin{equation}\label{eq:model0}
\vspace{-.2in}\end{equation}
\vspace{-.4in}\end{tcolorbox}
\end{center}
When $\f^{(i)}$'s are identical we also write $A(\d,\f)$ instead of $A(\d,\f^{(i)})$.

\noindent \textbf{Memory capacity:} One of the most fundamental features of any neural net (including single neurons as special cases) is their ability to properly store/memorize a large amount of data. To see how this can be achieved within the above NN mathematical formalism, assume first that one is given $m$ data pairs $(\x^{(0,k)},\y^{(0,k)})$, $k\in\{1,2,\dots,m\}$, where $\x^{(0,k)}\in \mR^{n}$ are $n$-dimensional data vectors and $\y^{(0,k)}\in\mR$ are their associated labels. Finding weights $W^{(i)}$  such that one in the above formalism obtains
\begin{equation}\label{eq:model3}
\x^{(1)}=\x^{(0,k)}\quad \Longrightarrow \quad \x^{(L+1)}=\y^{(0,k)} \qquad \forall k,
\end{equation}
is then sufficient to properly associate each of the data vectors with its corresponding label. For a given architecture $A(\d,\f^{(i)})$, the \emph{memory capacity, $C(A(\d,\f^{(i)}))$, } is then defined as the largest sample size, $m$, such that (\ref{eq:model3}) holds for any collection of data pairs $(\x^{(0,k)},\y^{(0,k)})$, $k\in\{1,2,\dots,m\}$ with certain prescribed properties. Given the importance of the role that the memory capacity plays in the overall mosaic of properly understanding the neural nets' functioning, we below present several results that, to a large degree, almost fully characterize it.

To make the analysis that follows easier to present, a few structural and technical assumptions are in place as well. Since these are fairly aligned with the ones discussed in \cite{Stojnictcmspnncaprdt23}, we here only briefly mention them and refer for a more detailed exposition to \cite{Stojnictcmspnncaprdt23}.

\noindent \textbf{Structural (network architecture) assumptions:} We consider 1-hidden layer zero-thresholds \emph{committee machines} sign perceptrons neural networks (SPNNs) which means the following: \emph{\textbf{(i)}} We assume $L=3$, $\b^{(i)}=0$, $W^{(1)}=I_{n\times n}$, and $W^{(3)}=\1_{d_{2}\times 1}^T$ (i.e. $W^{(3)}\in\mR^{1\times d_2}$ is a $d_2$-dimensional row vector of all ones). \textbf{\emph{(ii)}} We consider \emph{identity} neuronal functions in the first layer and zero-thresholds sign perceptrons in the hidden layers and at the output, i.e., we take $\f^{(1)}(\x^{(1)})=\x^{(1)}$ and $\f^{(i)}(W^{(i)}\x^{(i)}-\b^{(i)})=\mbox{sign} \left ( W^{(i)}\x^{(i)}\right )$ for $i=\{2,3\}$. \emph{\textbf{(iii)}}  We define $d\triangleq d_2$ and $\delta\triangleq \delta_1=\frac{d_1}{d_2}=\frac{n}{d}$ and, to make the main concepts easier to present, we assume that $d$ is any (odd) natural number.

When $W^{(i)}$ is a full matrix, the above architecture corresponds to the so-called \emph{fully connected committee machines} (FCM). On the other hand, if $W^{(i)}$ has a particular sparse structure, where the support of its $j$-th row, $\mbox{supp}\lp W_{j,:}^{(i)}\rp$, satisfies
$\mbox{supp}\lp W_{j,:}^{(i)}\rp=\cS^{(j)}$, with $\cS^{(j)}\triangleq\{(j-1)\delta+1,(j-1)\delta+2,\dots,j\delta\}$, then the above architecture corresponds to the so-called \emph{treelike committee machines} (TCM).

\noindent \textbf{Technical (data related) assumptions:} \emph{\textbf{(i)}} We assume  \emph{binary} labeling $\y_i^{(0,k)}\in\{-1,1\}$ (choosing \emph{sign} perceptrons as neuronal functions naturally lends itself to the binary labeling choice as well). \emph{\textbf{(ii)}} Inseparable data sets are not allowed (for example, indistinguishable/contradictory pairs (or subgroups) like $(\x^{(0,k)},\y^{(0,k)})$ and $(\x^{(0,k)},-\y^{(0,k)})$ can not appear). \emph{\textbf{(iii)}} We assume statistical data sets and in particular take $\x^{(0,k)}$ as iid standard normals. This follows the statistical trend from the classical single perceptron references (see, e.g., \cite{DTbern,Gar88,StojnicGardGen13,Cover65,Winder,Winder61,Wendel62}) and allows for, what is expected to be, a fairly universal statistical treatment. It should also be noted, that when one is concerned with providing universal memory capacity upper bounds, choosing any type of acceptable data set actually suffices.


\noindent \textbf{Relevant prior work:} Due to a direct connection between the memory capacity of spherical perceptrons and several fundamental questions in integral geometry, the early capacity considerations effectively stretch back to some of the geometrical/probabilistic  classic works (see, e.g., \cite{Schlafli,Cover65,Wendel62,Joseph60}). Certainly, the most closely associated result with the capacity of the spherical \emph{sign} perceptrons is that it \emph{doubles the dimension} of the data ambient space, $n$, i.e., $C(A(1;\mbox{sign}))\rightarrow 2n$ as $n\rightarrow\infty$. This was initially obtained in \cite{Schlafli,Cover65,Winder,Winder61,Wendel62,Cameron60,Joseph60} and later rediscovered and reproved in various different forms in a host of scientific fields ranging from machine learning, pattern recognition, high-dimensional geometry to information theory, probability, and statistical physics  \cite{BalVen87,Ven86,DT,StojnicISIT2010binary,DonTan09Univ,DTbern,Gar88,StojnicGardGen13,StojnicGardSphErr13}.

\underline{\emph{Networks of perceptrons:}}All the above works effectively ensure that the properties of a \emph{single} \emph{sign} perceptron are fairly well understood. On the other hand, as one moves to the corresponding multi-perceptron counterparts, the existing results and overall understanding of the underlying phenomena do not seem as strong. The TCM architectures related results are particularly scarce. On the other hands, a bit more is known about the FCM ones. However, a direct connection between the two is not easy to establish. Apart from the trivial fact that the FCM capacities are upper-bounds on the corresponding TCM ones, one may also (particularly in 1-hidden layer architectures) view the TCM capacities as roughly the FCM ones divided by $d$. These interesting connections ensure that being aware of the known FCM results is rather useful. However, almost all known results seem to relate the memory capacity, in one form or the other, to the total number of the so-called free network parameters (weights), $w=\sum_{i=1}^{L-1} d_id_{i+1}$. In particular, a result that is quite likely the most closely related to our own is the VC-dimension \emph{qualitative} memory capacity upper bound $O(w\log(w))$ (for the 1-hidden layer NNs, $w=d_1d_{2}+d_2=(n+1)d$ for FCM and $w=d_1+d_{2}=n+d$ for TCM which, for large $d_i$'s and huge $n$, gives, the above mentioned, ``\emph{division by $d$}'' relation between the FCM and TCM capacities). While they are not directly related, we mention a couple of results regarding the corresponding lower bounds as well. It was argued in \cite{Baum88} that for a shallow 3-layer network (similar to the one that we study here) the capacity scales as $O(nd)$. A much stronger version was obtained recently in \cite{Vershynin20}, where, for the networks with more than three layers, the capacity is shown to be (roughly speaking) at least $O(w)$.

\underline{\emph{Different activating functions:}} The \emph{sign} perceptrons are clearly among the functionally simplest types of neurons. Yet, due to their discreteness, among the simple functional structures, they are probably the hardest to analytically handle. Deviating from discreteness and allowing for various well known continuous counterparts/relaxations (i.e., for $\f$'s being sigmoid, ReLU, tanh and so on) makes things a bit easier and, consequently, a little bit more is known about the capacities of such structures. In particular, it was suggested for deep nets in \cite{Yama93}, and proven for 4-layer nets in \cite{GBHuang03}, that the capacity is at least $O(w)$ for sigmoids. In \cite{ZBHRV17,HardrtMa16} similar results were shown for ReLU with an additional restriction on the number of nodes that was later on removed in \cite{YunSuJad19} for both ReLU and tanh.

\underline{\emph{Practical achievability:}} We should also mention another line of work that is not directly related to what we study here, but it gain a sizeable popularity over the last several years. Namely, when one relaxes things and instead of the discrete neuronal functions considers continuous ones, efficient  algorithms can be designed to potentially approach the capacity. A great work has been done recently in this direction with the main focus on showing that the simple gradient based methods might actually perform quite well in this context. In particular, a lot of effort was put forth to show that the so-called mild over-parametrization (moderately larger number of all free parameters, $w$, compared to the size of the memorizable data set, $m$) suffices to ensure excellent gradient based methods  performance. A whole lot on the recent progress in rigorously establishing these results can be found in e.g. \cite{DuZhaiPoc18,GeWangZhao19,ADHLW19,JiTel19,LiLiang18,OymSol19,RuoyuSun19,SongYang19,ZCZG18}. While a majority of these works relates to FCMs, they are also extendable to TCMs as well.

\underline{\emph{Replica theory (statistical physics):}} Replica tools from statistical physics are a rather helpful (and often, the only available) tool when one faces hard analytical problems. The problems of our interest here are a no exception and many excellent relevant results, obtained via replica methods, are available. We recall that the main hardness here is that we are interested in very \emph{precise} capacity characterizations as functions of the number of the hidden layer neurons $d$. That means that any \emph{qualitative/scaling} characterizations (say, of the $O(\cdot)$ type) are not admissible. With very few exceptions, almost all of the known, and the above mentioned, prior works do relate to the scaling type of the capacity behavior. On the other hand, the replica methods based ones are not and, instead consider very precise analyses. These are mathematically non-rigorous, but are most closely related to our work in terms of both the studied models and the obtained capacity predictions. For example, \cite{EKTVZ92,BHS92} studied the very same, TCM architecture (as well as a directly related, FCM one) and  obtained the closed form replica symmetry based capacity predictions for any number, $d$, of the neurons in the hidden layer. Moreover, they established the corresponding large $d$ scaling behavior and showed that it violates the one obtained through the uniform-bounding extension of \cite{Cover65,Winder,Winder61,Wendel62} given in \cite{MitchDurb89}. To remedy this contradiction, they proceeded by studying the first level of replica symmetry breaking (rsb) and showed that it promises to lower the capacity. Corresponding large $d$ scaling rsb considerations were presented in \cite{MonZech95} for both the committee and the so-called parity machines (PM) (more on the earlier PM replica considerations can be found in, e.g., \cite{BarKan91,BHK90}). Also, for the FCM architecture, a bit later, \cite{Urban97,XiongKwonOh97} obtained the large $d$ scaling that matches the upper-bounding one of \cite{MitchDurb89}. More recently, \cite{BalMalZech19} obtained the first level of rsb capacity predictions for the TCM architectures with the ReLU activations. On the other hand,  \cite{ZavPeh21} moved things even further and obtained similar predictions for several different activations, including quadratic, erf, linear, and ReLU.

\noindent \textbf{Our contributions:} Within the above statistical context, we study the so-called $n$-scaled memory capacity of 1-hidden layer TCM SPNNs, i.e. we study
\begin{equation}\label{eq:model4}
c(d;\mbox{sign})\triangleq\lim_{n\rightarrow\infty} \frac{C(A([n,d,1];\mbox{sign}))}{n}.
\end{equation}
A very strong progress in characterizing $C(A([n,d,1];\mbox{sign}))$ for any given (odd) $d$ has been made in \cite{Stojnictcmspnncaprdt23}. In particular, utilizing the powerful Random Duality Theory (RDT) mathematical engine, \cite{Stojnictcmspnncaprdt23} provides an explicit upper bound $\hat{c}(d;\mbox{sign})$ on $c(d;\mbox{sign})$. Numerical results obtained for smaller values of $d$ suggested a strong benefit in adding more neurons in a network architecture context. On the other hand, we, in this paper, make a substantial progress in several different aspects including both methodological and practical ones.

\begin{figure}[h]
\centering
\centerline{\includegraphics[width=0.7\linewidth]{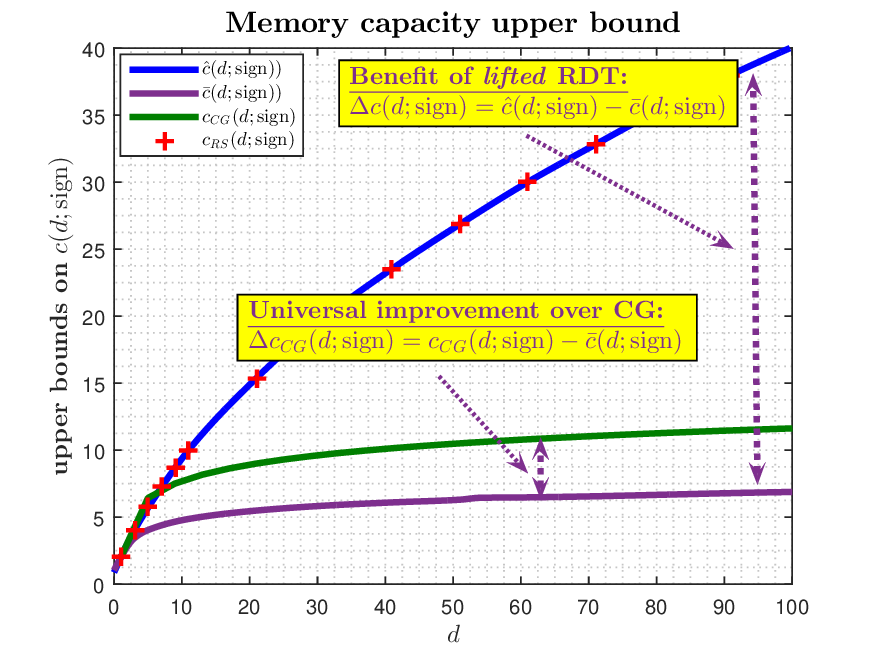}}
\caption{Memory capacity upper bound as a function of the number of neurons, $d$, in the hidden layer; 1-hidden layer TCM SPNN; \prp{\textbf{\emph{lifted} RDT}} versus  \bl{\textbf{\emph{plain}  RDT}}, \dgr{\textbf{Combinatorial geometry (CG)}}, and \red{\textbf{Replica symmetry (RS)}}}
\label{fig:fig1}
\end{figure}

\underline{\emph{A summary of the main technical results of the paper:}}  \emph{\textbf{(i)}} We first observe that for $\hat{c}(d;\mbox{sign})$ from \cite{Stojnictcmspnncaprdt23}, one has $\hat{c}(d;\mbox{sign})\sim \sqrt{d}$; we then rigorously show that $\lim_{d\rightarrow\infty}\frac{\hat{c}(d;\mbox{sign})}{\sqrt{d}}=6\sqrt{\frac{2}{\pi}}$. \emph{\textbf{(ii)}} We then observe that if one were to extrapolate large $d$ towards $n$, i.e., if one were to take $d\sim n$, the above formulas would give $C(A([n,d,1l;\mbox{sign}]))\sim n\sqrt{n}$. This, on the other hand, is a bit overly optimistic as the VC upper-bounding for $(L-2)$-hidden layer network in general gives $O(w\log(w))$ which for 1-hidden layer net ($L=3$) and $w\sim n$ would become $O(n\log(n))$. While this reasoning is not fully rigorous, it already hints that although the RDT produces what are expected to be excellent results for smaller $d$'s, it may overestimate a bit when it comes to the very large $d$. Above all, the scaling $\hat{c}(d;\mbox{sign})\sim \sqrt{d}$ directly violates the best known $\sim \log(d)$ one from \cite{MitchDurb89}. \emph{\textbf{(iii)}}  We then consider the recently developed \emph{fully lifted} RDT (fl RDT) as an alternative. Since the fl RDT typically relies on heavy numerical evaluations, we circumvent its full implementation and here focus on its a simplified, \emph{partially lifted}, variant and show that it allows for very neat, closed form, analytical capacity characterizations. Moreover, we obtain concrete bounds that \emph{universally} improve, for \emph{any} $d$, over the best known, mathematically rigorous, ones of \cite{MitchDurb89}. The obtained \emph{partially lifted} RDT results together with how they compare to the regular \emph{plain} RDT ones from \cite{Stojnictcmspnncaprdt23} are shown in Table \ref{tab:tab1} for a few smallest values of $d$ and in Figure \ref{fig:fig1} for a much wider range of $d$.

 \begin{table}[h]
  \caption{\textbf{\prp{Theoretical estimates}} of the memory capacity upper bounds of 1-hidden layer TCM SPNN}
  \label{tab:tab1}
  \centering
  \begin{tabular}{cccccc}
    \hline\hline
  \textbf{Upper bound on} & \textbf{Reference}  & \multicolumn{4}{c}{$d$}                   \\
    \cline{3-6}
    $c(d;\mbox{sign})\triangleq\lim_{n\rightarrow\infty} \frac{C(A([n,d,1];\mbox{sign}))}{n}$ \hspace{-.1in} &  (methodology)   & $\mathbf{1}$   & $\mathbf{3}$     & $\mathbf{5}$ & $\mathbf{7}$ \\
    \hline\hline
    $\bar{c}(d;\mbox{sign})$ & $ $  \hspace{.2in}   this paper \prp{(\textbf{\emph{lifted} RDT)}} \hspace{.2in} $ $   & \prp{$\mathbf{2}$} & \prp{$\mathbf{3.43}$}  & \prp{$\mathbf{4.03}$} & \prp{$\mathbf{4.39}$}     \\
\hline
   $ $  \hspace{.2in} $\hat{c}(d;\mbox{sign})$ \hspace{.2in} $ $ &  \cite{Stojnictcmspnncaprdt23} \bl{\textbf{(RDT)}}& \bl{$\mathbf{2}$} & \bl{$\mathbf{4.02}$}  & \bl{$\mathbf{5.77}$}& \bl{$\mathbf{7.31}$}  \\
    \hline
    $c_{RS}(d;\mbox{sign})$ & $ $  \hspace{.0in}   \cite{EKTVZ92,BHS92} \red{(\textbf{Replica symmetry)}} \hspace{.0in} $ $   & \red{$\mathbf{2}$} & \red{$\mathbf{4.02}$}  & \red{$\mathbf{5.77}$} & \red{$\mathbf{7.31}$}      \\
     \hline
    $c_{CG}(d;\mbox{sign})$ & $ $  \hspace{.0in}   \cite{MitchDurb89} \dgr{(\textbf{Combinatorial geometry)}}  \hspace{.0in} $ $   & \dgr{$\mathbf{2}$} & \dgr{$\mathbf{5.42}$}  & \dgr{$\mathbf{6.43}$} & \dgr{$\mathbf{7.05}$}  \\
       \hline\hline
  \end{tabular}
\end{table}

\section{Asymptotic RDT analysis}
\label{sec:analysis}

We start by revisiting the results of \cite{Stojnictcmspnncaprdt23} and reassessing their closeness to optimality by showing their precise large $d$ behavior. To that end, we recall that, after setting $W\triangleq W^{(2)}$, $W^{(1)}=I$, and $W^{(3)}=\1^T$ and
\begin{equation}\label{eq:ta6}
\y\triangleq \begin{bmatrix}
               \y^{(0,1)} & \y^{(0,2)} & \dots & \y^{(0,m)}
            \end{bmatrix}^T   \qquad \mbox{and} \qquad X\triangleq \begin{bmatrix}
               \x^{(0,1)} & \x^{(0,2)} & \dots & \x^{(0,m)}
            \end{bmatrix}^T,
\end{equation}
\cite{Stojnictcmspnncaprdt23} proceeded and obtained the following
\begin{center}
 	\tcbset{beamer,sidebyside,lower separated=false, fonttitle=\bfseries, coltext=black,
		interior style={top color=yellow!20!white, bottom color=yellow!60!white},title style={left color=black, right color=red!50!blue!60!white},
		width=(\linewidth-4pt)/4,before=,after=\hfill,fonttitle=\bfseries,equal height group=AT}
 	\begin{tcolorbox}[title=Algebraic memorization characterization of 1-hidden layer TCM SPNN:,sidebyside,width=1\linewidth]
\vspace{-.15in}\begin{eqnarray}\label{eq:ta8}
\hspace{-.3in} 0=\xi\triangleq \min_{W,Q} & & \hspace{-.1in}\|\y-\mbox{sign}(\mbox{sign}(Q) \1)\|_2 \nonumber \\
\hspace{-.5in} \mbox{subject to} & & \hspace{-.1in}XW^T=Q
\end{eqnarray}
 \tcblower
 \hspace{-.2in}$\Longleftrightarrow$ \hspace{.1in} Data set $\left (X,\y \right )$ is properly memorized.
 		\vspace{-.0in}
 	\end{tcolorbox}
\end{center}
\cite{Stojnictcmspnncaprdt23} then went further and analyzed the above optimization via RDT machinery and also obtained the following characterization of the upper bound on the $n$-scaled memory capacity
\begin{eqnarray}\label{eq:aan0}
\hspace{-.0in} \hat{c}(d;\mbox{sign})=\frac{d}{\frac{1}{2^d}\sum_{l=1}^{\lceil \frac{d}{2}\rceil } \binom{d}{\lceil\frac{d}{2}\rceil -l} \varphi_1(l;d)}, \quad \mbox{where} \quad  \varphi_1(l;d)    \triangleq  \mE_{G}\left \|\bar{G}_{1:l}^{(\lfloor\frac{d}{2}\rfloor+l)}\right \|_2^2,
\end{eqnarray}
and $\bar{G}^{(p)}$ is the vector obtained by taking the first $p$ components of $G_{i,1:d}$ (comprised of $d$ iid standard normals) and sorting them in the increasing order of their magnitudes. Moreover, \cite{Stojnictcmspnncaprdt23} continued even further and precisely determined $\varphi_1(l;d)$ for any $l$ and $d$. Since we are interested in the asymptotic large $d$ behavior in this section, the analysis of \cite{Stojnictcmspnncaprdt23} might not necessarily be the most adequate route to follow. We here adopt a different approach and, instead, follow the ideas presented in \cite{StojnicCSetam09}. To that end, we consider $\g\in\mR^{(\lfloor\frac{d}{2}\rfloor+l)}$ that has the iid standard normal components and observe that
\begin{equation}\label{eq:aan1}
  \varphi_1(l;d)  \triangleq  \mE_{G} \min_{\a_i\in\{0,1\},\|\a\|_1=l}  \sum_{i=1}^{(\lfloor\frac{d}{2}\rfloor+l)}\g_i^2\a_i.
  \end{equation}
Writing Lagrangian then gives
\begin{equation}\label{eq:aan2}
  \varphi_1(l;d)  \geq  \mE_{G} \max_{\nu}\min_{\a_i\in\{0,1\}}  \sum_{i=1}^{(\lfloor\frac{d}{2}\rfloor+l)}(\g_i^2-\nu)\a_i + \nu l
  =\mE_{G} \max_{\nu}  \sum_{i=1}^{(\lfloor\frac{d}{2}\rfloor+l)}\min(\g_i^2-\nu,0)+ \nu l.
  \end{equation}
After writing the integrals one further has
\begin{equation}\label{eq:aan3}
  \varphi_1(l;d)  \geq   \max_{\nu}
   \sum_{i=1}^{(\lfloor\frac{d}{2}\rfloor+l)} \frac{2}{\sqrt{2\pi}}\int_{0}^{\sqrt{\nu}}(\g_i^2-\nu)e^{-\frac{\g_i^2}{2}}d\g_i+ \nu l.
  \end{equation}
Assuming that $\nu\ll 1 $ (below we double check if this choice indeed makes sense), we have
{\small\begin{eqnarray}\label{eq:aan4}
  \varphi_1(l;d)  & \geq  &  \max_{\nu}
   \sum_{i=1}^{\lfloor\frac{d}{2}\rfloor+l} \frac{2}{\sqrt{2\pi}}\int_{0}^{\sqrt{\nu}}(\g_i^2-\nu)e^{-\frac{\g_i^2}{2}}d\g_i+ \nu l   \nonumber \\
   & \approx &  \max_{\nu}
   \sum_{i=1}^{\lfloor\frac{d}{2}\rfloor+l} \frac{2}{\sqrt{2\pi}}\int_{0}^{\sqrt{\nu}}(\g_i^2-\nu)\lp 1-\frac{\g_i^2}{2}\rp d\g_i+ \nu l \approx    \max_{\nu}
   -\lp \left \lfloor\frac{d}{2}\right \rfloor +l \rp \frac{4}{3\sqrt{2\pi}}\nu^{\frac{3}{2}}  + \nu l.\nonumber \\
  \end{eqnarray}}
Taking derivative gives
\begin{equation}\label{eq:aan5}
   -\lp \left \lfloor\frac{d}{2}\right \rfloor +l \rp \frac{2}{\sqrt{2\pi}}\nu^{\frac{1}{2}}  + l=0 \quad \Longleftrightarrow \quad
   \nu^{\frac{1}{2}} =\frac{l}{\lp \left \lfloor\frac{d}{2}\right \rfloor +l \rp \frac{2}{\sqrt{2\pi}}}.
   \end{equation}
A combination of (\ref{eq:aan4}) and (\ref{eq:aan5}) finally gives
\begin{equation}\label{eq:aan6}
 \varphi_1(l;d)   \geq    \frac{\pi}{6}\frac{l^3}{\lp \left \lfloor\frac{d}{2}\right \rfloor +l\rp^2}.
   \end{equation}
We now recall the following generic approximation (valid as long as $|n-2r|=o(n^{1/2})$) of the binomial coefficients
\begin{equation}\label{eq:aan7}
\binom{n}{r} \sim \frac{2^n}{\sqrt{\frac{n\pi}{2}}}e^{-\frac{(n-2r)^2}{2n}}.
\end{equation}
Taking $n=d$ and $r=\left \lfloor \frac{d}{2} \right\rfloor +l$, we obtain from (\ref{eq:aan7})
\begin{equation}\label{eq:aan8}
\binom{d}{\left \lfloor \frac{d}{2} \right\rfloor +l} \sim \frac{2^d}{\sqrt{\frac{d\pi}{2}}}e^{-\frac{(2l)^2}{2d}}.
\end{equation}
Combining (\ref{eq:aan1}), (\ref{eq:aan7}), and (\ref{eq:aan8}) we find
\begin{eqnarray}\label{eq:aan9}
\hspace{-.0in} \hat{c}(d;\mbox{sign})=\frac{1}{\frac{1}{2^d}\sum_{l=1}^{\lceil \frac{d}{2}\rceil } \binom{d}{\lceil\frac{d}{2}\rceil -l} \varphi_1(l;d)}\leq \frac{1}{\frac{1}{2^d}\sum_{l=1}^{\lceil \frac{d}{2}\rceil } \frac{2^d}{\sqrt{\frac{d\pi}{2}}}e^{-\frac{(2l)^2}{2d}}
\frac{\pi}{6}\frac{l^3}{\lp \left \lfloor\frac{d}{2}\right \rfloor +l\rp^2}}.
\end{eqnarray}
To get the appropriate scaling, we take $l=c\sqrt{\frac{d}{2}}$ with $c$ independent of $d$ (in a combination with (\ref{eq:aan5}) this choice ensures that indeed $\nu\ll 1$). We then obtain from (\ref{eq:aan8})
\begin{eqnarray}\label{eq:aan10}
\hspace{-.0in} \hat{c}(d;\mbox{sign}) \leq \frac{1}{\frac{1}{2^d}\sum_{l=1}^{\lceil \frac{d}{2}\rceil } \frac{2^d}{\sqrt{\frac{d\pi}{2}}}e^{-\frac{(2l)^2}{2d}}
\frac{\pi}{6}\frac{l^3}{\lp \left \lfloor\frac{d}{2}\right \rfloor +l\rp^2}}
\longrightarrow \frac{6}{\pi}\sqrt{d}\frac{1}{\sqrt{\frac{2}{\pi}}\int_{0}^{\infty}c^3e^{-c^2}dc}=6\sqrt{\frac{2}{\pi}}\sqrt{d}.
\end{eqnarray}
The above results are summarized in the following theorem.
\begin{theorem}(Memory capacity RDT based upper bound; large $d$ asymptotic) Consider 1-hidden layer TCM SPNN with architecture $A([n,d,1];\mbox{sign})$, standard normal iid data $X$, and $n$-scaled capacity $c(d;\mbox{sign})$  defined in (\ref{eq:model4}). As proven in \cite{Stojnictcmspnncaprdt23}, for
$\hat{c}(d;\mbox{sign})$  from (\ref{eq:aan0}), $\lim_{n\rightarrow\infty}\mP_{X}(c(d,\mbox{sign})<\hat{c}(d,\mbox{sign}))\longrightarrow 1$. Moreover, one has the following:
 \vspace{-.04in}\begin{center}
\tcbset{beamer,lower separated=false, fonttitle=\bfseries,
coltext=black , interior style={top color=orange!10!yellow!30!white, bottom color=yellow!80!yellow!50!white}, title style={left color=orange!10!cyan!30!blue, right color=green!70!blue!20!black}}
 \begin{tcolorbox}[beamer,title=\textbf{Asymptotic in $d$ ($n$-scaled) memory capacity upper bound:},lower separated=false, fonttitle=\bfseries,width=.7\linewidth] 
\vspace{-.1in}
{\small\begin{equation}\label{eq:aan11}
\hspace{-.0in} \lim_{d\rightarrow\infty}\frac{c(d;\mbox{sign})}{\sqrt{d}} \leq \lim_{d\rightarrow\infty}\frac{\hat{c}(d;\mbox{sign})}{\sqrt{d}} = \mathbf{6\sqrt{\frac{2}{\pi}}}\approx \mathbf{4.79}. \end{equation}}
\vspace{-.15in}
 \end{tcolorbox}
\end{center}\vspace{-.0in}
  \label{thm:thm1}
\end{theorem}\vspace{-.13in}
\begin{proof}
Follows from the above discussion.\end{proof}
To follow typical scholar presentation, inequalities are used in all of the above derivations. The underlying convexities and concentrations ensure that everything actually holds with equality, exactly  as stated in (\ref{eq:aan11}).

The above theorem establishes a neat convenient asymptotic result. However, it turns out that its analytical strength does not match its visual elegance. Namely, a hypothetical  extrapolating of large $d$ towards $n$, i.e., taking $d\sim n$, in the above formulas would give $C(A([n,d,1l;\mbox{sign}]))\sim n\sqrt{n}$. This, however, is a bit optimistic as the VC based upper bounds in general give $O(w\log(w))$ (where $w$ is the number of free parameters) which for 1-hidden layer net ($L=3$), where $w\sim n$, becomes $O(n\log(n))$. While this reasoning lacks a full mathematical rigor, it is intuitively sufficient to hint that even though the RDT produces excellent results for smaller $d$'s, it may overestimate by a bit when it comes to the very large $d$'s. On top of this generic reasoning, one has for the particular 1-hidden layer TCM SPNN architecture, the rigorously known scaling $\sim \log(d)$ from \cite{MitchDurb89} which is directly violated by the above $\hat{c}(d;\mbox{sign})\sim \sqrt{d}$. In the following section, we introduce a mechanism that \emph{universally} improves over \cite{MitchDurb89}.

\section{\emph{Lifted} Random Duality Theory (RDT)}
\label{sec:liftedrdt}

To upper bound the memory capacity, \cite{Stojnictcmspnncaprdt23} conducted an RDT analysis of the memorization condition from (\ref{eq:ta8}). It
heavily relied on the powerful RDT concepts developed in a long series of work \cite{StojnicCSetam09,StojnicICASSP10var,StojnicICASSP10block,StojnicRegRndDlt10}. As the analysis of the previous section showed, while the results of \cite{Stojnictcmspnncaprdt23} made a strong progress for small $d$, they are still far away from the optimal ones over the entire range of $d$'s. The recent development of the \emph{fully lifted} (fl) RDT \cite{Stojnicsflgscompyx23,Stojnicnflgscompyx23,Stojnicflrdt23} is then naturally the way to proceed to remedy this problem. However, the fl RDT heavily relies on substantial numerical evaluations which are here additionally complicated by the internal structure of the underlying problems. Combining this with the fact that the expected $\sim\sqrt{\log(d)}$ large $d$ capacity behavior takes rather astronomical values of $d$ to be distinguishable, suggests that it may be practically beneficial to take a different, analytically less accurate but more convenient route. Namely, we consider a \emph{partially lifted} (pl) RDT variant for which we obtain elegant closed form solutions. The pl RDT relies on the following principles.
\vspace{-.0in}\begin{center}
 	\tcbset{beamer,lower separated=false, fonttitle=\bfseries, coltext=black ,
		interior style={top color=yellow!20!white, bottom color=yellow!60!white},title style={left color=black!80!purple!60!cyan, right color=yellow!80!white},
		width=(\linewidth-4pt)/4,before=,after=\hfill,fonttitle=\bfseries}
 \begin{tcolorbox}[beamer,title={\small Summary of the \emph{partially lifted} (pl) RDT's main principles} \cite{StojnicCSetam09,StojnicRegRndDlt10,StojnicLiftStrSec13,StojnicGardSphErr13,StojnicGardSphNeg13}, width=1\linewidth]
\vspace{-.15in}
{\small \begin{eqnarray*}
 \begin{array}{ll}
\hspace{-.19in} \mbox{1) \emph{Finding underlying optimization algebraic representation}}
 & \hspace{-.0in} \mbox{2) \emph{Determining the \textbf{partially lifted} random dual}} \\
\hspace{-.19in} \mbox{3) \emph{Handling the \textbf{partialy lifted} random dual}} &
 \hspace{-.0in} \mbox{4) \emph{Double-checking strong random duality.}}
 \end{array}
  \end{eqnarray*}}
\vspace{-.25in}
 \end{tcolorbox}
\end{center}\vspace{-.0in}

We assume a complete familiarity with both the plain RDT and the pl RDT and below discuss each of the above four principles within the context of our interest here.

\vspace{.1in}
\noindent \underline{1) \textbf{\emph{Algebraic memorization characterization:}}}  The following lemma, proven in \cite{Stojnictcmspnncaprdt23}, provides a convenient optimization representation of the memorization property.
\begin{lemma}(\cite{Stojnictcmspnncaprdt23} Algebraic optimization representation)
Assume a 1-hidden layer TCM SPNN with architecture $A([n,d,1];\mbox{sign})$. Any given data set $\left (\x^{(0,k)},1\right )_{k=1:m}$ can not be properly memorized by the network if
\begin{equation}\label{eq:ta10}
  f_{rp}(X)>0,
\end{equation}
where
\begin{equation}\label{eq:ta11}
f_{rp}(X)\triangleq \frac{1}{\sqrt{n}}\min_{\|\z^{(j)}\|_2=1,Q} \max_{\Lambda\in\mR^{m\times d}} \|\1-\emph{\mbox{sign}}(\emph{\mbox{sign}}(Q) \1)\|_2 +\sum_{j=1}^{d}(\Lambda_{:,j})^TX^{(j)}\z^{(j)} -\tr(\Lambda^TQ),
\end{equation}
and $X\triangleq \begin{bmatrix}
               \x^{(0,1)} & \x^{(0,2)} & \dots & \x^{(0,m)}
            \end{bmatrix}^T$.
  \label{lemma:lemma1}
\end{lemma}

In what follows, we consider mathematically the most challenging, so-called \emph{linear}, regime with
\begin{equation}\label{eq:ta14}
  \alpha\triangleq \lim_{n\rightarrow\infty}\frac{m}{n}.
\end{equation}
As observed in \cite{Stojnictcmspnncaprdt23}, the above lemma holds for any given data set $\left (\x^{(0,k)},1\right )_{k=1:m}$. On the other hand, to analyze (\ref{eq:ta10}) and (\ref{eq:ta11}), the RDT proceeds by imposing a statistics  on $X$.


\noindent \underline{2) \textbf{\emph{Determining the lifted random dual:}}} We first recall on the  utilization of the so-called concentration of measure property, which basically means that for any fixed $\epsilon >0$,  we have (see, e.g. \cite{StojnicCSetam09,StojnicRegRndDlt10,StojnicICASSP10var})
\begin{equation*}
\lim_{n\rightarrow\infty}\mP_X\left (\frac{|f_{rp}(X)-\mE_X(f_{rp}(X))|}{\mE_X(f_{rp}(X))}>\epsilon\right )\longrightarrow 0.\label{eq:ta15}
\end{equation*}
Another key ingredient of the RDT machinery is the following so-called \emph{partially lifted} random dual theorem.
\begin{theorem}(Memorization characterization via \emph{partially lifted} random dual) Let $d$ be any odd integer. Consider a TCM SPNN with $d$ neurons in the hidden layer and architecture $A([n,d,1];\mbox{sign})$, and let the elements of $X\in\mR^{m\times n}$ , $G\in\mR^{m\times d}$, and $H\in\mR^{\delta\times d}$ be iid standard normals. Assuming $c_3>0$, set
\vspace{-.0in}
\begin{eqnarray}
\phi(Q) & \triangleq & \|\1-\emph{\mbox{sign}}(\emph{\mbox{sign}}(Q) \1)\|_2\nonumber \\
 f_{rd}^{(1)}(G) & \triangleq &  \max_{\phi(Q)=0}  -c_3  \|G-Q\|_F  \nonumber \\
 f_{rd}^{(2)}(H) & \triangleq &  \|H\|_F  \nonumber \\
 \bar{\phi}_0(\alpha;c_3)  & \triangleq & \lim_{n\rightarrow\infty} \frac{1}{\sqrt{n}}\lp  \frac{c_3}{2}
- \frac{1}{c_3}\log \lp \mE_{G}e^{\frac{c_3}{2}f_{rd}^{(1)}(G)}\rp
- \frac{1}{c_3}\log \lp \mE_{H}e^{\frac{c_3}{2}f_{rd}^{(2)}(H)}\rp   \rp.\label{eq:ta16}
\vspace{-.04in}\end{eqnarray}
One then has \vspace{-.02in}
\begin{eqnarray}
\hspace{-.1in}(\bar{\phi}_0(\alpha;c_3)  > 0)   &  \Longrightarrow  &  \lp \lim_{n\rightarrow\infty}\mP_{X}(f_{rp}>0)\longrightarrow 1 \rp  \nonumber \\
& \Longrightarrow & \lp \lim_{n\rightarrow\infty}\mP_{X}(A([n,d,1];\mbox{sign}) \quad \mbox{fails to memorize data set} \quad (X,\1))\longrightarrow 1\rp.\label{eq:ta17}
\end{eqnarray}
\label{thm:thm2}
\end{theorem}\vspace{-.17in}
\begin{proof}
    A complete familiarity with the basics of RDT from \cite{StojnicCSetam09,StojnicICASSP10block,StojnicICASSP10var,StojnicRegRndDlt10,StojnicGardSphErr13} and the main novelties are discussed. To that end, we start by considering a bounded function $\phi(Q)<\bar{c}_1$ and for $\bar{c}\gg\bar{c}_1$ and $X\in\mR^{m\times n}$ , $\Lambda,Q,G\in\mR^{m\times d}$, $H\in\mR^{\delta\times d}$, $\z^{(j)}\in\mR^{\delta\times 1}$, $\g'\in\mR^{j\times 1}$,  we set:
\begin{eqnarray}\label{eq:supp1}
\xi_{rp}(X) & \triangleq & \frac{1}{\sqrt{n}} \lp \min_{\|\z^{(j)}\|_2=1,Q} \max_{\|\Lambda\|_F=\bar{c}} \lp \phi(Q) +\sum_{j=1}^{d}(\Lambda_{:,j})^TX^{(j)}\z^{(j)} -\tr(\Lambda^TQ)\rp +\sum_{j=1}^{d}\|\Lambda_{:,j}\|_2\g_j' \rp \nonumber \\
 \xi_{rd}(G,H) & \triangleq & \frac{1}{\sqrt{n}}
  \min_{\|\z^{(j)}\|_2=1,Q}\max_{\|\Lambda\|_F=\bar{c}}\lp \phi(Q)+ \tr(\Lambda^T G)+\sum_{j=1}^{d}\|\Lambda_{:,j}\|_2(H_{:,j})^T\z^{(j)} -\tr(\Lambda^TQ) \rp.
\end{eqnarray}
For iid standard normals $X$, $G$, $H$ , and $g'$, \cite{Stojnictcmspnncaprdt23} showed that the following generic result holds:
\begin{equation}
\mP_X (\xi_{rp}(X)>0)\geq \mP_{G,H} (\xi_{rd}(G,H)>0).\label{eq:supp2}
\end{equation}
Particularizing to $\phi(Q)=0$ and $\bar{c}=1$ and utilizing the lifting machinery (see, e.g., \cite{Gordon85,StojnicGardSphErr13,StojnicMoreSophHopBnds10}), one then has even stronger for $c_3> 0$
\begin{multline}
\mE_{X,g'} \mbox{exp}\lp \max_{\|\z^{(j)}\|_2=1,\phi(Q)=0} \min_{\|\Lambda\|_F=1} c_3\lp \sum_{j=1}^{d}(\Lambda_{:,j})^TX^{(j)}\z^{(j)}  -\tr(\Lambda^TQ)\rp +c_3 g'\rp\leq \\
\mE_{G,H} \mbox{exp}\lp  \max_{\|\z^{(j)}\|_2=1,\phi(Q)=0} \min_{\|\Lambda\|_F=1} c_3\lp \tr(\Lambda^T G)+\sum_{j=1}^{d}\|\Lambda_{:,j}\|_2(H_{:,j})^T\z^{(j)} -\tr(\Lambda^TQ) \rp
\rp,\label{eq:supp3}
\end{multline}
where $g'$ is standard normal independent of all other random quantities. Applying $\log$ on both sides gives
\begin{multline}
\frac{c_3^2}{2}+\log\lp\mE_{X,g'} \mbox{exp}\lp \max_{\|\z^{(j)}\|_2=1,\phi(Q)=0} \min_{\|\Lambda\|_F=1} c_3\lp \sum_{j=1}^{d}(\Lambda_{:,j})^TX^{(j)}\z^{(j)} -\tr(\Lambda^TQ)\rp \rp \rp \leq \\
\log\lp\mE_{G,H} \mbox{exp}\lp  \max_{\|\z^{(j)}\|_2=1,\phi(Q)=0} \min_{\|\Lambda\|_F=1} c_3\lp \tr(\Lambda^T G)+\sum_{j=1}^{d}\|\Lambda_{:,j}\|_2(H_{:,j})^T\z^{(j)} -\tr(\Lambda^TQ)\rp
\rp\rp.\label{eq:supp4}
\end{multline}
A bit of additional algebraic transformations further gives
\begin{multline}
\frac{c_3^2}{2}-\log\lp\mE_{G,H} \mbox{exp}\lp  \max_{\|\z^{(j)}\|_2=1,\phi(Q)=0} \min_{\|\Lambda\|_F=1} c_3\lp \tr(\Lambda^T G)+\sum_{j=1}^{d}\|\Lambda_{:,j}\|_2(H_{:,j})^T\z^{(j)} -\tr(\Lambda^TQ)\rp
\rp\rp
\leq \\
-\log\lp\mE_{X,g'} \mbox{exp}\lp \max_{\|\z^{(j)}\|_2=1,\phi(Q)=0} \min_{\|\Lambda\|_F=1} c_3\lp \sum_{j=1}^{d}(\Lambda_{:,j})^TX^{(j)}\z^{(j)} -\tr(\Lambda^TQ)\rp \rp \rp.\label{eq:supp5}
\end{multline}
Getting $\log$ inside the expectation on the right hand side and scaling everything by $c_3$ first gives
\begin{multline}
\frac{c_3}{2}-\frac{1}{c_3}\log\lp\mE_{G,H} \mbox{exp}\lp  \max_{\|\z^{(j)}\|_2=1,\phi(Q)=0} \min_{\|\Lambda\|_F=1} c_3\lp \tr(\Lambda^T G)+\sum_{j=1}^{d}\|\Lambda_{:,j}\|_2(H_{:,j})^T\z^{(j)} -\tr(\Lambda^TQ) \rp
\rp\rp
\leq \\
- \mE_{X} \lp \max_{\|\z^{(j)}\|_2=1,\phi(Q)=0} \min_{\|\Lambda\|_F=1} \lp \sum_{j=1}^{d}(\Lambda_{:,j})^TX^{(j)}\z^{(j)} -\tr(\Lambda^TQ)\rp \rp.
\label{eq:supp6}
\end{multline}
One then trivially also has
\begin{multline}
\frac{c_3}{2}-\frac{1}{c_3}\log\lp\mE_{G,H} \mbox{exp}\lp  \max_{\|\z^{(j)}\|_2=1,\phi(Q)=0} \min_{\|\Lambda\|_F=1} c_3\lp \tr(\Lambda^T G)+\sum_{j=1}^{d}\|\Lambda_{:,j}\|_2(H_{:,j})^T\z^{(j)} -\tr(\Lambda^TQ) \rp
\rp\rp
\leq \\
 \mE_{X} \lp \min_{\|\z^{(j)}\|_2=1,\phi(Q)=0} \max_{\|\Lambda\|_F=1} \lp \sum_{j=1}^{d}(\Lambda_{:,j})^TX^{(j)}\z^{(j)} -\tr(\Lambda^TQ)\rp \rp.
\label{eq:supp6a0a0}
\end{multline}
Solving the inner minimization over $\Lambda$ and maximization over $\z^{(j)}$ on the left hand side, we further have
\begin{multline}
\frac{c_3}{2}-\frac{1}{c_3}\log\lp\mE_{G,H} \mbox{exp}\lp  \max_{\phi(Q)=0} \min_{\|\Lambda\|_F=1} c_3\lp \sum_{j=1}^{d}\|\Lambda_{:,j}\|_2\lp -\|G_{:,j}-Q_{:,j}\|_2 +\|H_{:,j})\|_2 \rp \rp
\rp\rp
\leq \\
 \mE_{X} \lp \min_{\|\z^{(j)}\|_2=1,\phi(Q)=0} \max_{\|\Lambda\|_F=1} \lp \sum_{j=1}^{d}(\Lambda_{:,j})^TX^{(j)}\z^{(j)} -\tr(\Lambda^TQ)\rp \rp.
\label{eq:supp6a0}
\end{multline}
Optimizing further the left hand side over $\|\Lambda_{:,j}\|_2$ gives
\begin{multline}
\frac{c_3}{2}-\frac{1}{c_3}\log\lp\mE_{G,H} \mbox{exp}\lp  \max_{\phi(Q)=0}  -c_3 \sqrt{ \sum_{j=1}^{d}\lp -\|G_{:,j}-Q_{:,j}\|_2 +\|H_{:,j})\|_2 \rp^2 }
\rp\rp
\leq \\
 \mE_{X} \lp \min_{\|\z^{(j)}\|_2=1,\phi(Q)=0} \max_{\|\Lambda\|_F=1} \lp \sum_{j=1}^{d}(\Lambda_{:,j})^TX^{(j)}\z^{(j)} -\tr(\Lambda^TQ)\rp \rp.
\label{eq:supp6a1}
\end{multline}
After a few additional algebraic transformations, we also have
\begin{multline}
\frac{c_3}{2}-\frac{1}{c_3}\log\lp\mE_{G,H} \mbox{exp}\lp  \max_{\phi(Q)=0}  -c_3 \sqrt{\|G-Q\|_F^2 -2\sum_{j=1}^{d}  \|G_{:,j}-Q_{:,j}\|_2 \|H_{:,j})\|_2 +\|H\|_F^2  }
\rp\rp
\leq \\
 \mE_{X} \lp \min_{\|\z^{(j)}\|_2=1,\phi(Q)=0} \max_{\|\Lambda\|_F=1} \lp \sum_{j=1}^{d}(\Lambda_{:,j})^TX^{(j)}\z^{(j)} -\tr(\Lambda^TQ)\rp \rp.
\label{eq:supp6a2}
\end{multline}
By the Cauchy-Schwartz inequality, we find
\begin{equation}
\sum_{j=1}^{d}  \|G_{:,j}-Q_{:,j}\|_2 \|H_{:,j})\|_2
\leq \sqrt{\sum_{j=1}^{d} \|G_{:,j}-Q_{:,j}\|_2^2} \sqrt{\sum_{j=1}^{d}\|H_{:,j})\|_2^2}
=\|G-Q\|_F\|H\|_F.
\label{eq:supp6a3}
\end{equation}
We then also easily find
\begin{eqnarray}
\sqrt{\|G-Q\|_F^2 -2\sum_{j=1}^{d}  \|G_{:,j}-Q_{:,j}\|_2 \|H_{:,j})\|_2 +\|H\|_F^2  }
& \geq &
\sqrt{\|G-Q\|_F^2 -2\|G-Q\|_F\|H\|_F +\|H\|_F^2  } \nonumber \\
& \geq & \|G-Q\|_F -\|H\|_F.
\label{eq:supp6a4}
\end{eqnarray}
Connecting (\ref{eq:supp6a2}) and (\ref{eq:supp6a4}), we obtain
\begin{multline}
\frac{c_3}{2}-\frac{1}{c_3}\log\lp\mE_{G,H} \mbox{exp}\lp  \max_{\phi(Q)=0}  -c_3 \lp \|G-Q\|_F -\|H\|_F \rp
\rp\rp
\leq \\
 \mE_{X} \lp \min_{\|\z^{(j)}\|_2=1,\phi(Q)=0} \max_{\|\Lambda\|_F=1} \lp \sum_{j=1}^{d}(\Lambda_{:,j})^TX^{(j)}\z^{(j)} -\tr(\Lambda^TQ)\rp \rp.
\label{eq:supp6a5}
\end{multline}
One then, finally, has
\begin{multline}
\frac{c_3}{2}-\frac{1}{c_3}\log\lp\mE_{G} \mbox{exp}\lp c_3 f_{rd}^{(1)}(G)\rp\rp-\frac{1}{c_3}\log\lp\mE_{H} \mbox{exp}\lp  c_3 f_{rd}^{(2)}(H)
\rp\rp
\leq \\
 \mE_{X} \lp \min_{\|\z^{(j)}\|_2=1,\phi(Q)=0} \max_{\|\Lambda\|_F=1} \lp \sum_{j=1}^{d}(\Lambda_{:,j})^TX^{(j)}\z^{(j)} -\tr(\Lambda^TQ)\rp \rp.
\label{eq:supp7}
\end{multline}
Keeping in mind (\ref{eq:ta11}) and the underlying concentrations, a comparison of (\ref{eq:supp7}) and (\ref{eq:ta16}) completes the proof.
 \end{proof}

%
%
%
%
%
\noindent \underline{3) \textbf{\emph{Handling the lifted random dual:}}} After the above handling of the optimizations over $\Lambda$ and $\z^{(j)}$, one proceeds with a detailed careful analysis of the optimization over $Q$  and arrives at the following theorem.
\begin{theorem}(Memory capacity partially lifted (pl) RDT based upper bound; general $d$) Assume the setup of Theorem \ref{thm:thm1} with general $d$. Let the network $n$-scaled capacity, $c(d,\mbox{sign})$, be as defined in (\ref{eq:model4}). First one has
 \begin{eqnarray}\label{eq:aan12}
 \bar{\varphi}_2(l;d) & \triangleq & l\binom{\lfloor\frac{d}{2}\rfloor+l}{l},  \quad \bar{\varphi}_4(g)\triangleq\mbox{erf}\lp\frac{|g|}{\sqrt{2}}\rp \nonumber \\
 \bar{\varphi}_1(l;d)  & = & \frac{ \bar{\varphi}_2(l;d) }{\sqrt{1+\frac{c_3}{2\gamma}}^{l-1}}
\frac{1}{\sqrt{2\pi}}\int_{-\infty}^{\infty}  \bar{\varphi}_4\lp |g|\sqrt{1+\frac{c_3}{2\gamma}}\rp^{l-1}\lp 1-\bar{\varphi}_4(|g|)\rp^{\left \lfloor \frac{d}{2} \right \rfloor}e^{-\frac{-\lp g\sqrt{1+\frac{c_3}{2\gamma}}\rp^2}{2}}dg
 \nonumber \\
I_Q & = & \frac{1}{2}+ \frac{1}{2^d}\sum_{l=1}^{\lceil \frac{d}{2}\rceil } \binom{d}{\lceil\frac{d}{2}\rceil -l} \bar{\varphi}_1(l;d) \nonumber \\
 I_{sph} & = & \gamma_{sph}-\frac{1}{2c_3}\log \lp 1-\frac{c_3}{2\gamma_{sph}}\rp, \quad  \gamma_{sph} =  \frac{c_3+\sqrt{c_3^2+4}}{4} \nonumber \\
\bar{\phi}_0(\alpha) &  = & \max_{c_3>0}\min_{\gamma} \lp \frac{c_3}{2} +\gamma -\frac{\alpha}{c_3}\log(I_Q)-I_{sph}\rp.
\end{eqnarray}
 Further, consider the following
\vspace{-.0in}
\vspace{-.0in}\begin{center}
\tcbset{beamer,lower separated=false, fonttitle=\bfseries,
coltext=black , interior style={top color=orange!10!yellow!30!white, bottom color=yellow!80!yellow!50!white}, title style={left color=orange!10!cyan!30!blue, right color=green!70!blue!20!black}}
 \begin{tcolorbox}[beamer,title=\textbf{($n$-scaled general $d$) memory capacity upper bound, $\bar{c}(d;\mbox{sign})$, that satisfies:},lower separated=false, fonttitle=\bfseries,width=.92\linewidth] 
\vspace{-.15in}
{\small\begin{eqnarray}\label{eq:aan13}
\bar{\phi}_0(\bar{c}(d;\mbox{sign}))=0 \quad \Longleftrightarrow \quad  \max_{c_3>0}\min_{\gamma} \lp \frac{c_3}{2} +\gamma -\frac{\bar{c}(d;\mbox{sign})}{c_3}\log(I_Q)-I_{sph}\rp = 0. \end{eqnarray}}
\vspace{-.15in}
 \end{tcolorbox}
\end{center}\vspace{-.0in}
Then for any sample complexity $m$ such that $\alpha\triangleq \lim_{n\rightarrow\infty}\frac{m}{n}>\bar{c}(d;\mbox{sign})$
\begin{eqnarray}
 \lim_{n\rightarrow\infty}\mP_{X}(A([n,d,1];\mbox{sign}) \quad \mbox{fails to memorize data set} \quad (X,\1))\longrightarrow 1,\label{eq:ta36}
\end{eqnarray}
and
\begin{eqnarray}
 \lim_{n\rightarrow\infty}\mP_{X}(c(d,\mbox{sign})<\bar{c}(d,\mbox{sign}))\longrightarrow 1.\label{eq:ta37}
\end{eqnarray}
\label{thm:thm3}
\end{theorem}\vspace{-.17in}

\begin{proof}
We split the proof onto two parts: \textbf{\emph{(i)}} Handling $\frac{1}{c_3\sqrt{n}}\log\lp\mE_{H} \mbox{exp}\lp  c_3 f_{rd}^{(2)}(H)\rp \rp$; and \textbf{\emph{(ii)}} Handling of $\frac{1}{c_3\sqrt{n}}\log\lp\mE_{G} \mbox{exp}\lp c_3 f_{rd}^{(1)}(G)\rp\rp$.

\underline{\textbf{\textbf{\emph{(i)}} Handling $\frac{1}{c_3\sqrt{n}}\log\lp\mE_{H} \mbox{exp}\lp  c_3 f_{rd}^{(2)}(H)\rp \rp$:}} We first observe
\begin{equation}\label{eq:supp8}
I_{sph} \triangleq \frac{1}{c_3\sqrt{n}}\log\lp\mE_{H} \mbox{exp}\lp  c_3 f_{rd}^{(2)}(H)\rp \rp =
 \frac{1}{c_3\sqrt{n}}\log\lp\mE_{H} \mbox{exp}\lp  c_3 \|H^T\|_F\rp \rp.
\end{equation}
In \cite{StojnicMoreSophHopBnds10}, it was determined after appropriate scaling, $c_3\rightarrow c_3\sqrt{n}$, that
 \begin{equation}\label{eq:supp9}
  I_{sph}  =  \gamma_{sph}-\frac{1}{2c_3}\log \lp 1-\frac{c_3}{2\gamma_{sph}}\rp, \quad  \gamma_{sph} =  \frac{c_3+\sqrt{c_3^2+4}}{4}.
\end{equation}

\underline{\textbf{\textbf{\emph{(ii)}} Handling $\frac{1}{c_3\sqrt{n}}\log\lp\mE_{G} \mbox{exp}\lp  c_3 f_{rd}^{(1)}(G)\rp \rp$:}} We again start by  observing
\begin{equation}\label{eq:supp10}
\log(I_{Q}') \triangleq \frac{1}{c_3\sqrt{n}} \log \lp \mE_{G} \mbox{exp}\lp  c_3 f_{rd}^{(2)}(G)\rp\rp =
 \frac{1}{c_3\sqrt{n}} \log \lp\mE_{G} \mbox{exp}\lp  -c_3 \min_{\phi(Q)=0}\|G-Q\|_F\rp \rp.
\end{equation}
We now utilize the squaring trick introduced on a multitude of occasions in   \cite{StojnicGardSphErr13,StojnicMoreSophHopBnds10}
\begin{equation}\label{eq:supp11}
\log(I_{Q}')=\max_{\gamma}
 \frac{1}{c_3\sqrt{n}}\log \lp \mE_{G} \mbox{exp}\lp  c_3\lp -\frac{1}{4\gamma} \min_{\phi(Q)=0}\|G-Q\|_F^2 -\gamma  \rp\rp\rp.
\end{equation}
After  appropriate scaling, $c_3\rightarrow c_3\sqrt{n}$ and $\gamma\rightarrow \gamma\sqrt{n}$, and recalling $\alpha=\lim_{n\rightarrow\infty}\frac{m}{n}$, we further find
\begin{equation}\label{eq:supp12}
-\log(I_{Q}')=\min_{\gamma} \lp\gamma -
 \frac{\alpha}{c_3}\log \lp \mE_{G} \mbox{exp}\lp  c_3\lp -\frac{1}{4\gamma} \min_{\phi_(Q_{i,:})=1}\|G_{i,:}-Q_{i,:}\|_F^2 \rp\rp\rp\rp,
\end{equation}
where
\begin{equation}\label{eq:supp13}
  \phi_i(Q_{i,:})\triangleq \mbox{sign}(\mbox{sign}(Q_{i,:})\1).
\end{equation}
Utilizing the results of \cite{Stojnictcmspnncaprdt23}, we obtain
\begin{equation}\label{eq:supp14}
-\log(I_{Q}')=\min_{\gamma} \lp\gamma -
 \frac{\alpha}{c_3}\log \lp \frac{1}{2}+ \frac{1}{2^d}\sum_{l=1}^{\lceil \frac{d}{2}\rceil } \binom{d}{\lceil\frac{d}{2}\rceil -l}\mE_{G} \mbox{exp}\lp  c_3\lp -\frac{1}{4\gamma}  \|\bar{G}_{1:l}^{(\lfloor\frac{d}{2}\rfloor+l)} \|_2^2 \rp\rp\rp\rp,
\end{equation}
where $\bar{G}^{(p)}$ is the vector obtained by taking the first $p$ components of $G_{i,1:d}$ and sorting them in the increasing order of their magnitudes. We first set
\begin{equation}\label{eq:supp15}
 \bar{\varphi}_1(l;d)\triangleq \mE_{G} \mbox{exp}\lp  c_3\lp -\frac{1}{4\gamma}  \|\bar{G}_{1:l}^{(\lfloor\frac{d}{2}\rfloor+l)} \|_2^2 \rp\rp,
\end{equation}
and then take  a vector of  iid  standard normals, $\g\in\mR^{\lfloor\frac{d}{2}\rfloor+l}$,  to facilitate writing. Then assuming without a loss of generality that the first $l$ magnitudes of $\g$ are the smallest and  accounting for other symmetric scenarios via a combinatorial pre-factor, one has
   \begin{eqnarray}\label{eq:supp16}
\bar{\varphi}_1(l;d)  & = & \binom{\lfloor\frac{d}{2}\rfloor+l}{l}\lp \frac{1}{\sqrt{2\pi}}\rp^{\lfloor\frac{d}{2}\rfloor+l}\int_{\g_{1:l}}e^{-\frac{c3\|\g_{1:l}\|_2^2}{4\gamma}}
\int_{|\g_{l+1:\lfloor\frac{d}{2}\rfloor+l}|\geq \max(|\g_{1:l}|)}e^{-\frac{\|\g\|_2^2}{2}}d\g \nonumber \\
& = &  \binom{\lfloor\frac{d}{2}\rfloor+l}{l}\lp \frac{1}{\sqrt{2\pi}}\rp^{l}\int_{\g_{1:l}}e^{-\frac{c3\|\g_{1:l}\|_2^2}{4\gamma}}
\lp 1-\varphi_4(\max(|\g_{1:l}|))\rp^{\lfloor\frac{d}{2}\rfloor}e^{-\frac{\|\g_{1:l}\|_2^2}{2}}d\g_{1:l}.\nonumber \\
 \end{eqnarray}
One can then continue and, again without a loss of generality, assume that the largest of the $l$ smallest magnitudes of $\g$ is $\g_l$. Then accounting for other options through another combinatorial pre-factor, we obtain
   \begin{eqnarray}\label{eq:supp17}
\bar{\varphi}_1(l;d)
& = &  \binom{\lfloor\frac{d}{2}\rfloor+l}{l}\lp \frac{1}{\sqrt{2\pi}}\rp^{l}\int_{\g_{1:l}}e^{-\frac{c3\|\g_{1:l}\|_2^2}{4\gamma}}
\lp 1-\bar{\varphi}_4(\max(|\g_{1:l}|))\rp^{\lfloor\frac{d}{2}\rfloor}e^{-\frac{\|\g_{1:l}\|_2^2}{2}}d\g_{1:l}\nonumber \\
& = &  l\binom{\lfloor\frac{d}{2}\rfloor+l}{l}\lp \frac{1}{\sqrt{2\pi}}\rp^{l}\int_{\g_{l}}
\lp 1-\bar{\varphi}_4(|\g_{l}|)\rp^{\lfloor\frac{d}{2}\rfloor}e^{\frac{-\g_{l}^2}{2}} \nonumber \\
& & \times \int_{|\g_{1:l-1}|\leq |\g_l|}e^{-\frac{c3\|\g_{1:l}\|_2^2}{4\gamma}}e^{-\frac{\|\g_{1:l-1}\|_2^2}{2}}d\g_{1:l-1} \nonumber \\
& = &  \frac{l\binom{\lfloor\frac{d}{2}\rfloor+l}{l}}{\sqrt{1+\frac{c_3}{2\gamma}}^{l-1}} \frac{1}{\sqrt{2\pi}}\int_{\g_{l}}
\lp 1-\bar{\varphi}_4(|\g_{l}|)\rp^{\lfloor\frac{d}{2}\rfloor}e^{\frac{-\g_{l}^2(1+\frac{c_3}{2\gamma})}{2}}
\bar{\varphi}_4\lp|\g_{l}|\sqrt{1+\frac{c_3}{2\gamma}}\rp^{l-1} d\g_l.\nonumber \\
 \end{eqnarray}
The first combinatorial pre-factor $\binom{\lfloor\frac{d}{2}\rfloor+l}{l}$ accounts for the number of different smallest $l$ components  locations. The second  combinatorial pre-factor, $l$, accounts for the number of different choices for the location of the largest component  among the fixed smallest $l$ ones. Combining (\ref{eq:supp7}), (\ref{eq:supp8}), (\ref{eq:supp9}), (\ref{eq:supp10}), (\ref{eq:supp14}), and (\ref{eq:supp17}) and comparing to (\ref{eq:aan12}) completes the proof of Theorem \ref{thm:thm3}.
 \end{proof}


\noindent \underline{4) \textbf{\emph{Double checking the strong random duality:}}}  As already mentioned in \cite{Stojnictcmspnncaprdt23}, a standard double checking for the strong random duality is not in place as the typical, convexity based, considerations are inapplicable.

One can also analyze the behavior of the results given in the above theorem for large $d$. They scale as $\sim \log(d)$ which matches the scaling obtained in \cite{MitchDurb89} through the combinatorial geometry considerations of \cite{Cover65,Winder,Winder61,Wendel62}. However, such scaling needs rather astronomical values of $d$ to become relevant and is of not much practical use. This is in a sharp contrast with the $\sim\sqrt{d}$ behavior of the replica symmetry of \cite{EKTVZ92,BHS92}) and the plain RDT of \cite{Stojnictcmspnncaprdt23}, where even moderately large values of $d\sim 10000$ are sufficient to observe the precise constants of the scaling behavior. We therefore skip discussing the details of the large $d$ scaling analysis. Instead, we reemphasize the importance of Figure \ref{fig:fig1}. Namely, Figure \ref{fig:fig1} shows precisely for \emph{\textbf{any}} given (odd) $d$ how the obtained results compare to the ones of \cite{Stojnictcmspnncaprdt23} (and therefore the replica symmetry ones from \cite{EKTVZ92,BHS92}) and the previously best known, mathematically rigorous, ones of \cite{MitchDurb89}. One can clearly see that the partially lifted analysis is extremely beneficial.

\section{Conclusion}
\label{sec:conc}

In this paper we studied the treelike committee machines (TCM) \emph{sign} perceptron neural networks (SPNNs) and their memory capabilities. Utilizing a powerful mathematical machinery called Random Duality Theory (RDT), \cite{Stojnictcmspnncaprdt23} established a generic framework for the analysis of TCM SPNNs and made a very strong progress towards obtaining, in a mathematically rigorous way, their \emph{exact} capacities. A quick comparison of the results of \cite{Stojnictcmspnncaprdt23} and the classical corresponding single perceptron ones from \cite{Schlafli,Cover65,Winder,Winder61,Wendel62,Cameron60,Joseph60,BalVen87,Ven86,DT,StojnicISIT2010binary,DonTan09Univ,DTbern,Gar88,StojnicGardGen13,StojnicGardSphErr13},
leaves an  indication that a solid memorizing benefit can be expected from adding neurons in network configurations.

Since the results of \cite{Stojnictcmspnncaprdt23} are of the upper-bounding type, it is natural to wonder how far off from the optimal ones, they actually are. Here, we made a strong progress in this direction. To do so, we started by trying to get a feeling as to what kind of behavior one can expect for very large values of added neurons $d$.  We first conducted a mathematically rigorous analysis and obtained a very precise $d$-asymptotic estimates for the non-asymptotic results of \cite{Stojnictcmspnncaprdt23}. As they seemed to suggest a slight overestimation when compared to the known VC and combinatorial geometry based bounds, we pursued a route different from \cite{Stojnictcmspnncaprdt23}, and  utilized the \emph{partially lifted} Random Duality Theory (RDT) to get more accurate predictions. We first designed a generic analysis framework,  applicable for \textbf{\emph{any}} given (odd) number of the neurons in the hidden layer, and then showed that the \emph{partially lifted} RDT mechanism produces results \emph{universally} better than both the \emph{plain} RDT ones from \cite{Stojnictcmspnncaprdt23} and the previously best known ones of \cite{MitchDurb89}.

Many extensions are possible as well. The first next one is to conduct the analysis with the \emph{fully lifted} (fl) RDT (see, e.g., \cite{Stojnicflrdt23}). Also, in addition to the \emph{sign} activation functions considered here, many others are of interest including  ReLU, sigmoid, tanh, erf, quadratic, and so on. Different network architectures are of interest as well. These include the TCM ones with many instead of one layer as well as many forms of the FCM and PM ones. We will discuss all of these extensions in separate papers.

\begin{singlespace}
\bibliographystyle{plain}
\bibliography{nflgscompyxRefs}
\end{singlespace}

\end{document}